\documentclass[runningheads]{llncs}

\usepackage[T1]{fontenc}
\usepackage{amsmath,amssymb}
\usepackage{booktabs}
\usepackage{graphicx}
\usepackage{tikz}
\usetikzlibrary{arrows.meta, positioning, calc, patterns, decorations.pathreplacing,
  fit, backgrounds, decorations.markings, shapes.geometric, fadings, shadows}
\usepackage{pgfplots}
\pgfplotsset{compat=1.18}
\usepackage{etoolbox}
\AtBeginEnvironment{thebibliography}{\fontsize{8.5pt}{9.5pt}\selectfont%
  \setlength{\itemsep}{-1pt}\setlength{\parsep}{0pt}}

\newcommand{\Bel}{\mathrm{Bel}}
\newcommand{\Pl}{\mathrm{Pl}}

\begin{document}

\title{Neural Networks with Differentiable Belief-Function
Layers\thanks{Code: \texttt{https://github.com/sulcantonin/enn-public}}}
\titlerunning{Neural Networks with Differentiable Belief-Function Layers}

\author{Antonin Sulc\inst{1}\orcidID{0000-0001-7767-778X}}
\authorrunning{A. Sulc}
\institute{\email{sulc.antonin@gmail.com}}

\maketitle

\begin{abstract}
Yager's 1987 rule of combination $\boxplus$ was proposed on epistemic grounds: under high conflict, mass should be redistributed to ignorance rather than renormalised onto the class simplex. We observe that backpropagation needs the same rule: Dempster's normaliser $(1{-}\kappa)^{-1}$, where $\kappa$ is the conflict mass, puts a $(1{-}\kappa)^{-2}$ term in the gradient that diverges exactly when evidence sources disagree, while Yager's $\boxplus$ is bilinear and has no such term. Building on this, we construct a deep evidential pipeline in which every mass in the chain from heads to output is a valid Dempster--Shafer mass function, combined by Yager's rule. The output mass on the ignorance set, $m(\Theta)$, serves as the network's ``I do not know'' signal. The network is fit to hard labels and never sees an annotator distribution, yet $m(\Theta)$ orders human disagreement in two modalities, visual and linguistic. Dempster's rises too and ranks contested inputs about as well, but its normaliser divides the conflict out: Yager's vacuity is $9\times$ larger on the most contested CIFAR-10H images, so only Yager's is readable as an amount of ignorance without recalibration. For out-of-distribution detection it is competitive with strong single-pass baselines, while the clamped Dempster variant that scores higher becomes gradient-unstable once the clamp is removed. 

\end{abstract}

% ===================================================================
\section{Introduction}
\label{sec:intro}
% ===================================================================

Standard neural classifiers conflate ignorance with low probability,
because softmax normalises mass across all classes regardless of
whether the network has evidence for any of them.
Dempster--Shafer theory (DST)~\cite{shafer1976mathematical} resolves this by
letting the ignorance set $\Theta$ carry its own mass. The question we
take up is not
whether DST is the right framework but whether it can be implemented
inside a back-propagated network without losing DST semantics or
training stability. Belief-function layers have in fact been built on
learned features~\cite{denoeux2000neural,denoeux2019logistic,tong2021evidential},
but always with Dempster's rule, and two obstructions have kept the
construction awkward. First, classical DST is set-theoretic, with up to
$2^K$ focal elements for $K$ classes, and to keep enumeration feasible
neurosymbolic work often replaces exact $\Bel/\Pl$ with real-valued
surrogates~\cite{riegel2020logical}, which weaken the duality
$\Pl(A){=}1{-}\Bel(\bar A)$ (we write $\bar A=\Theta\setminus A$ for
the complement). Second, Dempster's rule of combination normalises by
$1/(1{-}\kappa)$, where $\kappa$ is the conflict mass, and its
derivative contains a $(1{-}\kappa)^{-2}$ term that destabilises
optimisation in exactly the high-conflict regime SGD produces.

Any model that fuses evidence from several sources must decide what
happens to the mass on which those sources disagree. The observation
that motivates this paper is that the rule long advocated on epistemic
grounds for that choice is the same rule back-propagation requires.
Yager's $\boxplus$, proposed in 1987 to answer conflict with ignorance
rather than amplified agreement~\cite{yager1987dempster}, is bilinear in the
input masses; with one argument fixed it is $1$-Lipschitz in $\ell_1$
(Theorem~\ref{thm:yager_stability}), so gradients through it cannot
amplify. Two communities four decades apart reach the same rule from
one shared root: the absence of the $1/(1{-}\kappa)$ normaliser. With
that rule in place the remaining obstructions go away: a small focal set
$\mathcal{F}=\{\{c_i\},\Theta\}$ keeps enumeration tractable and
$\Bel/\Pl$ exact, and Shafer's
$\alpha$-discounting~\cite{shafer1976mathematical} generalises from an
analyst-supplied scalar to a learnable per-source coefficient
$\alpha_s\in[0,1]$ conditioned on inter-head agreement.

Our contributions are: (i)~the observation that Yager's epistemic
argument and SGD's bounded-gradient requirement specify the same rule
(Sect.~\ref{sec:theory}); (ii)~an end-to-end neural DST stack whose
output vacuity separates absent from conflicting evidence by
construction (Sect.~\ref{sec:method}); and
(iii)~evidence that $m(\Theta)$ tracks human visual and linguistic
ambiguity (Sect.~\ref{sec:exp:human}).

% ===================================================================
\section{Related Work}
\label{sec:related}
% ===================================================================

Combination rules have been a central concern in DST since Zadeh's
critique of Dempster's normalisation under high
conflict~\cite{zadeh1986simple}; alternative
rules are surveyed in~\cite{sentz2002combination}. Yager's
rule~\cite{yager1987dempster} is one well-known response. To
our knowledge, the present paper is the first direct comparison
of Yager and Dempster as the combination layer of a back-propagated
network. We follow Subjective
Logic~\cite{josang2016subjective,vasilakes2025logic} for the vacuity
reading of $m(\Theta)$.

Belief functions inside classifiers have a long line of work, from the
evidential $k$-NN rule~\cite{denoeux1995knn} and neural
classifier~\cite{denoeux2000neural} to the DS-layer reading of logistic
regression~\cite{denoeux2019logistic} and evidential deep classifiers
that convert learned features into mass
functions~\cite{tong2021evidential,tong2021fusion}. These aggregate with
Dempster's rule; our question is what happens to that choice once the
aggregation has to be differentiated. Credal and random-set
networks~\cite{manchingal2025rsnn,wang2024credal,caprio2024credal}
pursue set-valued prediction by other means.

On the deep-learning side, single-pass methods (SNGP, DUQ,
EDL~\cite{liu2020simple,vanamersfoort2020duq,sensoy2018evidential},
Posterior Networks~\cite{charpentier2020posterior}) and sampling
alternatives (MC Dropout, Deep
Ensembles~\cite{gal2016dropout,lakshminarayanan2017simple}) summarise
predictive uncertainty, mostly as a single scalar; standard OOD
baselines (MSP~\cite{hendrycks2017baseline},
ODIN~\cite{liang2018enhancing}, Energy~\cite{liu2020energy}) are
collected in OpenOOD~\cite{yang2022openood}. EDL and PostNet are the
closest neighbours: both emit Dirichlet parameters and connect them to
DST at a single layer, without explicit multi-source combination, and
neither yields a $[\Bel,\Pl]$ interval.

% ===================================================================
\section{Method}
\label{sec:method}
% ===================================================================

We work throughout with the focal set
$\mathcal{F} = \{\{c_1\},\dots,\{c_K\},\Theta\}$ on a frame
$\Theta = \{c_1,\dots,c_K\}$: the smallest extension of the class set
that admits an explicit ignorance state, with $|\mathcal{F}| = K+1$
instead of $2^K$. A mass function whose every focal element is a
singleton or the whole frame is called
\emph{quasi-Bayesian}~\cite{daniel2012quasibayesian}; ours puts
positive mass on all $K{+}1$
elements of $\mathcal{F}$. This is not a general belief function on
$2^\Theta$, and we claim no such generality. We call the
resulting model the Mass-Primitive Evidential Neural Network (MP-ENN).

\subsection{Overview of the Forward Pass}
\label{sec:method:synthesis}

% Pipeline diagram palette
\definecolor{headA}{HTML}{2563EB}
\definecolor{fused}{HTML}{F59E0B}
\definecolor{panel}{HTML}{F8FAFC}
\definecolor{edgegray}{HTML}{64748B}
\definecolor{pipelabel}{HTML}{475569}

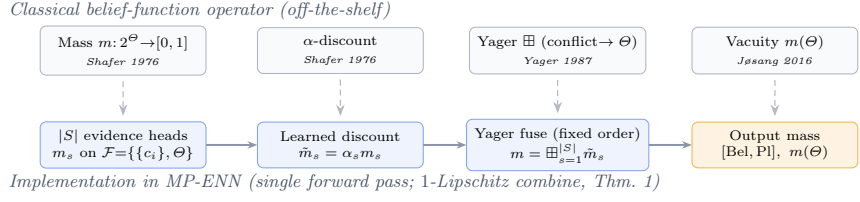
\begin{figure}[t]
\centering
\resizebox{0.92\textwidth}{!}{%
\begin{tikzpicture}[
    >=Stealth,
    every node/.style={inner sep=0pt, outer sep=0pt},
    shelfbox/.style={rounded corners=3pt, draw=edgegray!55, fill=panel,
                     align=center, inner sep=4pt, font=\scriptsize,
                     minimum width=2.9cm, minimum height=0.8cm},
    pipebox/.style={rounded corners=3pt, draw=headA!60, fill=headA!8,
                    align=center, inner sep=4pt, font=\scriptsize,
                    minimum width=2.9cm, minimum height=0.8cm},
    outbox/.style={rounded corners=3pt, draw=fused!80, fill=fused!12,
                   align=center, inner sep=4pt, font=\scriptsize,
                   minimum width=2.9cm, minimum height=0.8cm},
    pipearrow/.style={-{Stealth[length=5pt]}, line width=1pt, edgegray!75},
    plumb/.style={->, densely dashed, edgegray!50, line width=0.6pt,
                  shorten >=3pt, shorten <=3pt},
    sectionlabel/.style={font=\footnotesize\itshape, text=pipelabel,
                         anchor=west},
]
\node[sectionlabel] at (-0.4, 1.75)
    {Classical belief-function operator (off-the-shelf)};
\node[shelfbox] (mfn)   at (1.6, 1.05) {Mass $m{:}\,2^\Theta{\to}[0,1]$\\[1pt]\textit{\tiny Shafer 1976}};
\node[shelfbox] (alpha) at (5.4, 1.05) {$\alpha$-discount\\[1pt]\textit{\tiny Shafer 1976}};
\node[shelfbox] (yager) at (9.2, 1.05) {Yager $\boxplus$ (conflict$\to\Theta$)\\[1pt]\textit{\tiny Yager 1987}};
\node[shelfbox] (vac)   at (13.0,1.05) {Vacuity $m(\Theta)$\\[1pt]\textit{\tiny J{\o}sang 2016}};

\node[sectionlabel] at (-0.4, -1.25)
    {Implementation in MP-ENN (single forward pass; $1$-Lipschitz combine, Thm.~\ref{thm:yager_stability})};
\node[pipebox] (head) at (1.6,-0.6) {$|S|$ evidence heads\\$m_s$ on $\mathcal{F}{=}\{\{c_i\},\Theta\}$};
\node[pipebox] (disc) at (5.4,-0.6) {Learned discount\\$\tilde m_s=\alpha_s m_s$};
\node[pipebox] (combine) at (9.2,-0.6) {Yager fuse (fixed order)\\$m=\boxplus_{s=1}^{|S|}\tilde m_s$};
\node[outbox] (out) at (13.0,-0.6) {Output mass\\$[\Bel,\Pl],\ m(\Theta)$};

\draw[pipearrow] (head) -- (disc);
\draw[pipearrow] (disc) -- (combine);
\draw[pipearrow] (combine) -- (out);
\draw[plumb] (mfn) -- (head);
\draw[plumb] (alpha) -- (disc);
\draw[plumb] (yager) -- (combine);
\draw[plumb] (vac) -- (out);
\end{tikzpicture}%
}
\caption{MP-ENN forward pass. Each box below is implemented by the
classical operator directly above it; forward and backward both stay
inside DST.}
\label{fig:pipeline}
\end{figure}

Figure~\ref{fig:pipeline} shows the forward pass, which composes only
off-the-shelf operators: Shafer's mass function and
$\alpha$-discount~\cite{shafer1976mathematical}, Yager's
combination~\cite{yager1987dempster}, and J{\o}sang's vacuity reading of
$m(\Theta)$~\cite{josang2016subjective}. The chain rule passes through
every layer unchanged, so the same $m(\Theta)$ that J{\o}sang reads as
vacuity is what we later use as the OOD score and the
human-disagreement tracker
(Sect.~\ref{sec:exp:human}, Sect.~\ref{sec:exp:ood}): one quantity, not three.

\subsection{Mass-Primitive Evidence Heads}
\label{sec:method:heads}

A shared encoder $f_\theta$ (ResNet-18) produces features
$\mathbf{h} = f_\theta(\mathbf{x})$. The encoder is an ordinary network
and we make no belief-function claim about its activations; the DST
construction begins at the evidence heads and holds from there to the
output. A set $S$ of evidence heads
emits non-negative evidence
$e_{i,s} = \mathrm{softplus}(\mathbf{w}_{i,s}^\top \mathbf{h}{+}b_{i,s})$
for class $i\in\{1,\dots,K\}$ and source $s\in S$, and assembles its
mass as
\begin{equation}
m_s(\{c_i\}) = e_{i,s}/(E_s{+}W), \qquad
m_s(\Theta) = W/(E_s{+}W),
\label{eq:evidence_to_mass}
\end{equation}
with $E_s = \sum_{i=1}^K e_{i,s}$ and $W>0$ a prior weight. The construction
is mass-primitive: $\Bel(\{c_i\}) = m(\{c_i\})$ and $\Pl(\{c_i\}) =
m(\{c_i\}) + m(\Theta)$, so on $\mathcal{F}$ every class shares the
same $[\Bel,\Pl]$ width $m(\Theta)$, so any method with class-varying
widths does not realise exact DST here. Eq.~\eqref{eq:evidence_to_mass} re-reads Sensoy's
parametrization~\cite{sensoy2018evidential} as a direct mass
assignment rather than a Dirichlet hyperparameter; equivalently it is
the multinomial form of Walley's imprecise Dirichlet
model~\cite{walley1996idm}. This step is not new, and we state it only
because it fixes the interface the combination layer consumes; the
contribution of this paper lies downstream of it. Here $W$ acts as
the Dirichlet prior strength: $E_s+W$ is the effective ``total
evidence'' (data plus prior pseudo-counts), and $m_s(\Theta)$ is the
mass left to ignorance when data evidence is weak. We use $W{=}K$
throughout (CIFAR: $W{=}10$, ChaosNLI: $W{=}3$).

\subsection{Yager Combination, and Why Not Dempster}
\label{sec:method:yager}

Dempster's rule combines two mass functions $m_1, m_2$ by intersecting
their focal supports and dividing by $1 - \kappa$, where
$\kappa = \sum_{B\cap C=\emptyset} m_1(B)\,m_2(C)$ is the conflict mass.
Writing $\tilde m(A) = \sum_{B\cap C = A} m_1(B) m_2(C)$ for the
unnormalised intersection, so that $m_{1\oplus 2}(A) =
\tilde m(A)/(1-\kappa)$, the quotient rule gives
\begin{equation}
\frac{\partial m_{1\oplus 2}(A)}{\partial m_1(B)}
= \frac{1}{1-\kappa}\,\frac{\partial \tilde m(A)}{\partial m_1(B)}
+ \frac{\tilde m(A)}{(1-\kappa)^{2}}\,\frac{\partial \kappa}{\partial m_1(B)} .
\label{eq:dempster_grad}
\end{equation}
Both terms grow without bound as the sources disagree
($\kappa \to 1$). On $\mathcal{F}$ this is not asymptotic: taking
$m_1, m_2$ concentrated on two different singletons with mass $p$ each,
the largest partial derivative in Eq.~\eqref{eq:dempster_grad} tracks
$(1{-}\kappa)^{-1}$ exactly, rising from $1.33$ at $\kappa{=}0.25$ to
$4999.2$ at $\kappa{=}0.9998$. For Yager's rule it is $1.00$ at every
conflict level, attaining the bound of
Theorem~\ref{thm:yager_stability} rather than merely respecting it.
Training Dempster therefore requires clamping $1-\kappa$ away from
zero, which is a numerical guard with no belief-function reading, and
it suppresses exactly the high-conflict regime the theory exists to
reason about.

Yager's rule keeps the same intersection, but reassigns conflict to
$\Theta$ instead of normalising: $m_{1 \boxplus 2}(A) = \tilde m(A)$ for
$A \ne \Theta,\emptyset$, and $m_{1 \boxplus 2}(\Theta) =
m_1(\Theta) m_2(\Theta) + \kappa$. The two terms are separately
readable: the product is mass no source committed, $\kappa$ is mass the
sources committed to incompatible classes, so output vacuity reports not
only how much is unknown but why. Dempster's normaliser mixes them.
On $\mathcal{F} = \{\{c_i\},\Theta\}$ this is closed-form: every
cross-singleton intersection $\{c_i\}\cap\{c_j\}$ with $i\ne j$ is
$\emptyset$, so the only focal pairs contributing to $\{c_i\}$ are
$(\{c_i\},\{c_i\})$, $(\{c_i\},\Theta)$, and $(\Theta,\{c_i\})$;
the cross-singleton products $m_1(\{c_i\})m_2(\{c_j\})$ for $i\ne j$
all flow into the conflict mass $\kappa$ and hence into
$m_{1\boxplus 2}(\Theta)$:
\begin{align}
m_{1 \boxplus 2}(\{c_i\}) &= m_1(\{c_i\})m_2(\{c_i\}) +
m_1(\{c_i\})m_2(\Theta) + m_1(\Theta)m_2(\{c_i\}),
\label{eq:yager_cf_focal}\\
m_{1 \boxplus 2}(\Theta) &= m_1(\Theta)m_2(\Theta) +
\textstyle\sum_{i \ne j} m_1(\{c_i\})m_2(\{c_j\}).
\label{eq:yager_cf_theta}
\end{align}

\subsection{Multi-Source Fusion with Learned Reliability}
\label{sec:method:fusion}

Each head after the first is discounted by a learned reliability
$\alpha_s \in [0,1]$; the first head is left undiscounted
($\alpha_1{=}1$) so that it acts as the reference frame of the
sequential combination. Discounting is
via $m^\alpha(F) = \alpha\, m(F)$ for $F \ne \Theta$ and
$m^\alpha(\Theta) = 1 - \alpha(1 - m(\Theta))$. On $\mathcal{F}$ this
is equivalent to Shafer's general $\alpha$-discount (which adds
$(1{-}\alpha)$ to $m(\Theta)$ on $2^\Theta$): the only focal elements
are the singletons and $\Theta$, so once the singletons are scaled by
$\alpha$, mass conservation determines the $\Theta$ component.
Let $\mathbf{e}_s = (e_{1,s},\dots,e_{K,s})\in
\mathbb{R}^K$ be the per-source evidence vector (the softplus output
before normalisation in Eq.~\eqref{eq:evidence_to_mass}) and
$\bar{\mathbf{e}} = |S|^{-1}\sum_{s\in S} \mathbf{e}_s$ the across-head
consensus. We set $\alpha_s = \sigma\!\big(g_\phi(
[\mathbf{e}_s;\bar{\mathbf{e}};\mathbf{e}_s-\bar{\mathbf{e}}])\big)$,
where $g_\phi$ is a small two-layer MLP and the sigmoid $\sigma$ keeps
$\alpha_s\in(0,1)$. Sect.~\ref{sec:exp:ood} reports that this gate does
not, in fact, do measurable work on a shared encoder.
Writing $\tilde m_s$ for the discounted mass of
source $s$, the heads are fused by a left fold in the fixed index
order $s = 1,\dots,|S|$,
\begin{equation}
m^{(1)} = \tilde m_1, \qquad
m^{(s)} = m^{(s-1)} \boxplus \tilde m_s \ \ (s = 2,\dots,|S|), \qquad
m = m^{(|S|)},
\label{eq:fold}
\end{equation}
so the $|S|$ sources are combined in $|S|{-}1$ pairwise steps. Because
$\boxplus$ is commutative but not associative
(Sect.~\ref{sec:theory}), the result depends on this order
only through which source is folded in last; we measure the size of
that dependence in Sect.~\ref{sec:exp:ood}.

\subsection{Training Objective}
\label{sec:method:loss}

For ground-truth class $y$:
\begin{equation}
\mathcal{L} = -\log(m(\{c_y\}) + \epsilon)
+ \lambda_{\text{unc}}\, m(\Theta)
+ \frac{\lambda_{\text{aux}}}{|S|} \sum_{s\in S}
[-\log(m_s(\{c_y\}) + \epsilon)].
\label{eq:total_loss}
\end{equation}
The first term encourages correct mass on $\{c_y\}$, the second
penalises vacuity on in-distribution data, and the third prevents
auxiliary heads from collapsing. The constant $\epsilon{=}10^{-8}$ is
a numerical floor; we use $\lambda_{\text{unc}}{=}0.1$ and
$\lambda_{\text{aux}}{=}0.1$ throughout. The task term scores the
belief on the true class, $m(\{c_y\})=\Bel(\{c_y\})$: once $m(\Theta)$
is fixed (which the $\lambda_{\text{unc}}$ term controls), this is a
proper scoring rule. We do not claim joint propriety over the full
$(K{+}1)$-simplex (see Limitations).

% ===================================================================
\section{Theoretical Properties}
\label{sec:theory}
% ===================================================================

The pipeline is \emph{invariant by construction}: for every parameter
value, the per-source masses, their discounted forms and the fused
output are valid mass functions on $\mathcal{F}$, with $\Bel \le \Pl$, exact duality
$\Pl(A) = 1 - \Bel(\bar A)$, and the usual boundary conditions. Nothing
is projected or repaired and no surrogate stands in for $\Bel/\Pl$,
which is what neurosymbolic relaxations~\cite{riegel2020logical} give up
to stay tractable. The identities are immediate; that they survive
training is not, and that is what the bound below secures.
Separately, $\boxplus$ is commutative but not
associative~\cite{yager1987dempster,sentz2002combination}: conflict
pulled into $m(\Theta)$ at an early step cannot re-emerge as evidence
against later sources. We therefore fuse in the fixed order
of~\eqref{eq:fold}; a symmetric variant averaging over all $|S|!$
orderings is also implemented, and Sect.~\ref{sec:exp:ood} measures how
little the choice matters in practice. The one property that carries the
argument is the following; its proof is short and given in the extended
version.

\begin{theorem}[Gradient Stability of Yager]
\label{thm:yager_stability}
For every $A,B \in \mathcal{F}\setminus\{\emptyset\}$ and $i \in \{1,2\}$,
$|\partial m_{1\boxplus 2}(A)/\partial m_i(B)| \le 1$. With one argument
fixed, the map $m_1 \mapsto m_{1\boxplus 2}$ is linear and
column-stochastic, hence $1$-Lipschitz in $\ell_1$ on the simplex of
mass functions on $\mathcal{F}$.
\end{theorem}

% ===================================================================
\section{Experiments}
\label{sec:experiments}
% ===================================================================

\subsection{Setup}
\label{sec:exp:setup}

We run ten methods $\times$ five seeds each. The visual backbone is
ResNet-18 on CIFAR-10, with MP-ENN using three evidence heads
($|S|{=}3$), prior weight $W{=}K$ (so $W{=}10$ on CIFAR-10, $W{=}3$ on
ChaosNLI), $\lambda_{\text{unc}}{=}\lambda_{\text{aux}}{=}0.1$,
trained with Adam (batch size $128$, cosine LR schedule, gradient norm
clipped at $1.0$) for $200$ epochs. The clip is applied identically to
all methods, and the norms reported in Sect.~\ref{sec:exp:ood} are
pre-clip, so the Yager/Dempster contrast is not a clipping artefact.
For Dempster
training we clamp $1{-}\kappa$ at $10^{-8}$; the clamp binds only as
$\kappa\to 1$ and is disabled in the stress check of
Sect.~\ref{sec:exp:ood}.
The Softmax baselines (MSP, Energy, ODIN) are post-hoc scores on
MP-ENN-Yager's pre-softplus head logits (hence the shared ID accuracy
and ECE in Table~\ref{tab:ood_results}); all other baselines are
trained independently. OOD evaluation follows
OpenOOD~\cite{yang2022openood}: near-OOD is CIFAR-100; far-OOD covers
SVHN, DTD, MNIST, Fashion-MNIST, Gaussian and Uniform noise. Human
alignment uses CIFAR-10H~\cite{peterson2019human} (50 annotators per
image) for visual ambiguity, and ChaosNLI~\cite{nie2020can} (100
annotators per pair, RoBERTa-base) for linguistic ambiguity. Neither
annotator distribution is used in training: both models are fit to hard
labels, so the alignment reported below is not supervised.
Five-seed quantities in Table~\ref{tab:ood_results} report mean $\pm$
std; the human-alignment results (Sect.~\ref{sec:exp:human}) are point
estimates from the representative seed.

\subsection{Vacuity Tracks Human Disagreement (the Interpretability Test)}
\label{sec:exp:human}

DST's interpretability claim is structural: $m(\Theta)$ means ``I lack
evidence,'' not ``the classes are equiprobable.'' If real, it should
localise where humans do not know what to answer. We test this on two
datasets recording per-example disagreement.

Note that Eq.~\eqref{eq:total_loss} trains on hard labels and its
$\lambda_{\text{unc}}$ term pushes $m(\Theta)$ down on in-distribution
data, which looks like the opposite of what we claim. The two are
compatible: that term is uniform across examples, so it sets the overall
level of vacuity, not its per-example ordering. The per-example
variation comes from elsewhere. An ambiguous input drives the heads
apart, that disagreement is $\kappa$, and
Eq.~\eqref{eq:yager_cf_theta} adds $\kappa$ to $m(\Theta)$; nothing in
the loss reaches that term directly. The $\lambda_{\text{unc}}$ sweep
below agrees. No annotator distribution is seen in training, so the
alignment reported here is zero-shot.

\begin{figure}[t]
\centering
\includegraphics[width=0.70\textwidth]{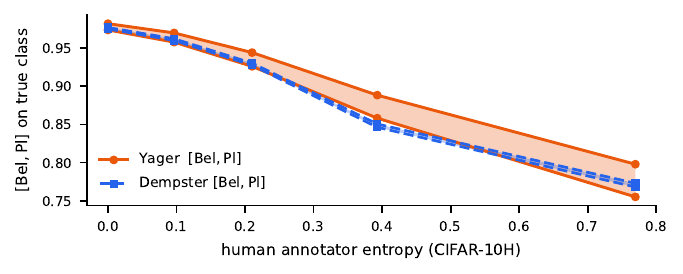}
\caption{Mean $[\Bel,\Pl]$ interval on the true class over the
$10{,}000$ CIFAR-10 test images, binned by CIFAR-10H annotator entropy.
Belief falls as annotators disagree, and under Yager the interval widens
with it ($0.008\to0.043$, a factor of $5$); under Dempster the band stays
hairline-thin ($0.002\to0.005$). The width is exactly $m(\Theta)$.}
\label{fig:human}
\end{figure}

\paragraph{CIFAR-10H (visual ambiguity).}
Across all 10{,}000 test images, Yager's vacuity rises with human
annotator entropy (Spearman $\rho{=}0.278$) \emph{monotonically} across
all five entropy bins, $0.008\to0.043$ (Fig.~\ref{fig:human}), and
separates the most- from the least-contested decile at AUROC $0.77$. It
is $9\times$ larger than Dempster's on that decile ($0.044$ vs.\
$0.005$). Dempster ranks those images almost as well
($\rho{=}0.263$, overlapping bootstrap intervals); what separates the
rules is scale, not ordering. Both use the same encoder architecture and
reach near-identical accuracy (Table~\ref{tab:ood_results}); the
difference is how much signal they leave: Yager adds the conflict mass
to ignorance, while Dempster's normaliser divides it out, so only
Yager's $m(\Theta)$ is readable on its own $[0,1]$ scale without
recalibration.

\paragraph{ChaosNLI (linguistic ambiguity).}
The same holds for language: on the $1{,}514$ ChaosNLI pairs
(RoBERTa-base, SNLI-trained to $91.7\%$ validation accuracy), vacuity
again rises with human disagreement (Spearman $\rho{=}0.346$,
$p<10^{-40}$), and mean vacuity is roughly twice as large on the most
contested quartile as on the least ($0.076$ vs.\ $0.036$).
Softmax entropy ($\rho{=}0.390$) and energy ($\rho{=}0.346$) match it as
scalars; the point is not that vacuity wins, but that one quantity
tracks both modalities and carries a $[\Bel,\Pl]$ interval scalars
cannot.

\subsection{OOD Detection (Secondary Capability)}
\label{sec:exp:ood}

The same $m(\Theta)$ is also a competitive single-pass OOD score
(Table~\ref{tab:ood_results}): clamped Dempster scores highest among
the single-pass methods (behind only the much costlier deep
ensemble), and Yager sits with energy scoring. Dempster's
trainability, however, rests on the $1{-}\kappa$ clamp---a numerical
guard with no DST reading. Over $200$ random mass pairs on $\mathcal{F}$
(mean conflict $\bar\kappa{=}0.75$) Dempster's gradients are seven times
Yager's ($\|\nabla\|=1.52$ vs.\ $0.21$): the $(1{-}\kappa)^{-2}$
blow-up of Sect.~\ref{sec:method:yager} made concrete. Dempster's edge
depends on clamping away exactly the high-conflict regime DST is meant
to reason about; Yager needs no guard.

\begin{table}[t]
\centering
\small
\setlength{\tabcolsep}{4pt}
\caption{Five-seed CIFAR-10/OpenOOD comparison; macro AUROC and FPR95
averaged across seven OOD datasets. Further five-seed AUROCs: ODIN
$0.871$, DUQ $0.859$, SNGP $0.886$, MC Dropout $0.885$.}
\label{tab:ood_results}
\begin{tabular}{@{}lcccc@{}}
\toprule
Method            & ID acc.\ (\%)  & ECE             & Macro AUROC          & Macro FPR95 \\
\midrule
Softmax + MSP     & $94.33{\pm}.13$ & $.036{\pm}.001$ & $0.653{\pm}.033$     & $0.853{\pm}.066$ \\
Softmax + Energy  & $94.33{\pm}.13$ & $.036{\pm}.001$ & $0.863{\pm}.056$     & $0.345{\pm}.079$ \\
EDL               & $94.04{\pm}.14$ & $.050{\pm}.003$ & $0.895{\pm}.031$     & $0.337{\pm}.072$ \\
MP-ENN-Yager      & $94.33{\pm}.13$ & $.036{\pm}.001$ & $0.860{\pm}.055$     & $0.342{\pm}.077$ \\
MP-ENN-Dempster   & $94.26{\pm}.14$ & $.042{\pm}.002$ & $0.912{\pm}.012$     & $0.292{\pm}.044$ \\
Deep Ensemble (5) & $94.20{\pm}.16$ & $.043{\pm}.002$ & $\mathbf{0.954}{\pm}.002$     & $\mathbf{0.164}{\pm}.005$ \\
\bottomrule
\end{tabular}
\end{table}

\paragraph{Ablations and compute.}
Prior weight $W\in\{1,5,10,20,100\}$ keeps near-OOD AUROC in
$[0.855, 0.866]$ and the ID/OOD vacuity gap essentially flat, so we
use $W{=}K{=}10$ on CIFAR. Regularisation
$\lambda_{\text{unc}}\in\{0,0.01,0.1,1\}$ moves AUROC from $0.865$ to
$0.850$. The learned discount is the one component we cannot claim
works. It sits at $\alpha_s\approx 1$ on undamaged heads, as it should,
but also under stress: with head $2$ fed Gaussian-corrupted features
($\sigma{=}1.5$, three seeds) it gives $\alpha = (0.99998, 0.99995,
0.99997)$. The corrupted head is ranked lowest in all three seeds, so
the gate identifies it, but a discount of $3\times 10^{-5}$ is
numerically inert. On a shared encoder the heads see identical features,
and inter-head disagreement appears too weak a signal to drive
discounting; independent sources would be the fair test.
Fusion order matters less than non-associativity might suggest:
commutativity collapses the $3!$ orderings to three distinct outcomes,
whose output masses differ by $0.002$ in $\ell_1$ on CIFAR-10 test with
no image changing class. On CIFAR-100, where mean conflict is $0.144$
rather than $0.022$, the spread grows to $0.012$ but vacuity AUROC still
varies by only $0.0004$. Inference is
single-pass at $51.9$\,ms, $14\%$ over Softmax+Energy.

% ===================================================================
\section{Discussion and Conclusion}
\label{sec:conclusion}
% ===================================================================

The argument rests on one observation: Yager's rule, advocated since
1987 because conflict should be reassigned to ignorance rather than
amplified into consensus, is also the rule a back-propagated network
needs, since dropping Dempster's $1/(1{-}\kappa)$ normaliser removes the
amplified consensus and the $(1{-}\kappa)^{-2}$ gradient blow-up at
once. The rest is small (polynomial focal set, learnable
$\alpha$-discount, chain rule), and the result runs Shafer's mass and
$\alpha$-discount, Yager's combination and J{\o}sang's vacuity with
exact $\Bel/\Pl$ in a single forward pass. The payoff is that
$m(\Theta)$ is readable at the output on its own scale, and tracks what
humans treat as ``I do not know'' without being trained to.

\emph{Limitations.} Focal elements are the singletons and $\Theta$
(quasi-Bayesian), so no mass can go to a subset such as $\{c_i,c_j\}$;
Yager combination is not associative; the task loss is conditionally
proper given $m(\Theta)$ but not jointly proper over the
$(K{+}1)$-simplex~\cite{gneiting2007strictly}, a known obstruction to
learning second-order uncertainty by loss
minimisation~\cite{bengs2022pitfalls}. Human-alignment numbers are
single-seed. The learned $\alpha$-discount does not survive its own
stress test (Sect.~\ref{sec:exp:ood}), so on a shared encoder it is
inactive rather than validated; making it work, most likely with
genuinely independent sources, is the natural follow-up, alongside
conformal prediction over $[\Bel,\Pl]$ intervals.

\begin{credits}
\subsubsection{\ackname}
Claude Code (Anthropic) was used to assist with brainstorming,
iterating on arguments, and implementing experiments.
\end{credits}

% Bibliography
\bibliographystyle{splncs04}
\bibliography{belief}

\end{document}